\documentclass[runningheads]{llncs}
\pdfoutput=1

\usepackage{amsmath,amssymb}
\usepackage{times,helvet,courier}
\usepackage{listings}
\newcommand{\clasp}{\textit{clasp}}

\newcommand{\gringo}{\textit{gringo}}

\newcommand{\sugar}{\textit{sugar}}
\newcommand{\aspartame}{\textit{aspartame}}

\newcommand{\Var}{\ensuremath{\mathcal{V}}}
\newcommand{\dom}[1]{\ensuremath{D(#1)}}
\newcommand{\Int}{\ensuremath{\mathcal{I}(\Var)}}
\newcommand{\Bool}{\ensuremath{\mathcal{B}(\Var)}}
\newcommand{\Con}{\ensuremath{\mathcal{C}}}
\newcommand{\opp}[1]{\ensuremath{\overline{#1}}}

\newcommand{\range}[1]{\ensuremath{I(#1)}}
\newcommand{\low}[1]{\ensuremath{\overrightarrow{l}(#1)}}
\newcommand{\upp}[1]{\ensuremath{\overrightarrow{u}(#1)}}
\newcommand{\blow}[1]{\ensuremath{\overleftarrow{l}(#1)}}
\newcommand{\bupp}[1]{\ensuremath{\overleftarrow{u}(#1)}}
\newcommand{\bound}[1]{\ensuremath{\ub(#1)}}
\newcommand{\ub}{\ensuremath{\mathit{ub}}}
\newcommand{\add}[1]{\ensuremath{S(#1)}}
\newcommand{\erg}[1]{\ensuremath{s(#1)}}

\newcommand{\ass}{\ensuremath{v}}
\newcommand{\val}[1]{\ensuremath{\ass(#1)}}

\newcommand{\diff}{\ensuremath{\mathit{alldifferent}}}

\newcommand{\code}[1]{\lstinline[basicstyle=\ttfamily]{#1}}

\newcommand{\SQUEEZE}{\!\!\!\!}

\title{Aspartame: Solving Constraint Satisfaction Problems with Answer Set Programming}

\author{M.~Banbara\inst{1}\! \and M.~Gebser\inst{2}\! \and K.~Inoue\inst{3}\! \and T.~Schaub\thanks{Affiliated with the
                              School of Computing Science at
                              Simon Fraser University,
                              Burnaby, Canada,
                              and the
                              Institute for Integrated and Intelligent Systems
                              at
                              Griffith University,
                              Brisbane, Australia.}\inst{2}\! \and T.~Soh\inst{1}\! \and N.~Tamura\inst{1}\! \and M.~Weise\inst{2}}
\institute{%
  University of Kobe \and University of Potsdam \and 
  National Institute of Informatics Tokyo} 

\titlerunning{Aspartame: Solving Constraint Satisfaction Problems with ASP}
\authorrunning{M.~Banbara \emph{et al}\/.} 

\begin{document}

\maketitle

\begin{abstract}
Encoding finite linear 
CSPs as Boolean formulas and
solving them by using modern SAT solvers has proven to be highly effective,
as exemplified by the award-winning \sugar\ system.
We here develop an alternative approach based on ASP.
This allows us to use first-order encodings providing us with a high degree of flexibility
for easy experimentation with different 
implementations. 
The resulting system \aspartame\ re-uses parts of \sugar\ for parsing and normalizing CSPs.
The obtained set of facts is then combined with an ASP encoding that can be grounded and solved by 
off-the-shelf ASP systems.
We establish the competitiveness of our approach by empirically contrasting \aspartame\ and  \sugar.
\end{abstract}


\section{Introduction}\label{sec:introduction}

Encoding finite linear Constraint Satisfaction Problems (CSPs; \cite{dechter03,CPHandbook})
as propositional formulas and 
solving them by using modern solvers for Satisfiability Testing
(SAT; \cite{SATHandbook}) has proven to be a highly effective approach,
as demonstrated by the award-winning \sugar\footnote{\url{\texttt{http://bach.istc.kobe-u.ac.jp/sugar}}} system.
The CSP solver \sugar\ reads a CSP instance and transforms it into a propositional formula in
Conjunctive Normal Form (CNF).
The translation relies on the order encoding \cite{crabak94a,tatakiba09a}, and
the resulting CNF formula can 
be solved by an off-the-shelf SAT solver.

In what follows,
we elaborate upon an alternative approach based on Answer Set Programming (ASP; \cite{baral02a}) 
and present the resulting CSP solver \aspartame\footnote{%
\url{\texttt{http://www.cs.uni-potsdam.de/wv/aspartame}}}.
The major difference between \sugar\ and \aspartame\ rests upon the implementation of the translation
of CSPs into Boolean constraint problems.
While
\sugar\ implements a translation into CNF in the imperative programming language JAVA,
\aspartame\
starts with a translation into a set of facts.\footnote{In practice, \aspartame\ re-uses \sugar's front-end for parsing and normalizing CSPs.}
In turn, these facts are combined with a general-purpose ASP encoding for CSP solving
(also based on the order encoding),
which is subsequently instantiated by an off-the-shelf ASP grounder.
The resulting propositional logic program is then solved by an off-the-shelf ASP solver.

The high-level approach of ASP has obvious advantages.
First, 
instantiation is done by general-purpose ASP grounders rather than dedicated implementations.
Second,
the elaboration tolerance of ASP allows for easy maintenance and modifications of encodings.
And finally,
it is easy to experiment with novel or heterogeneous encodings.
However, the intruding question is whether the high-level approach of \aspartame\ matches the
performance of the more dedicated \sugar\ system. 
We empirically address this question by contrasting the performance of both CSP solvers,
while fixing the back-end solver to \clasp, used as both a SAT and an ASP solver.

From an ASP perspective, we gain insights into advanced modeling techniques for solving CSPs.
The ASP encoding implementing 
CSP solving with \aspartame\ has the following features:
\begin{itemize}
\item usage of function terms to abbreviate structural subsums
\item avoidance of (artificial) intermediate Integer variables (
      to break 
      sum expressions)
\item order encoding applied to structural subsum variables (
as well as input variables)
\item encoding-wise filtering of relevant threshold values (no blind usage of 
domains)
\item customizable ``pigeon-hole constraint'' encoding for 
alldifferent constraints
\item ``smart'' encoding of table constraints, tracing admissible tuples along 
arguments 
\end{itemize}

In the sequel,
we assume some familiarity with ASP, its semantics as well as its basic language constructs.
A comprehensive treatment of ASP can be found in \cite{baral02a},
one oriented towards ASP solving is given in \cite{gekakasc12a}.
Our encodings are given in the language of \gringo~3\ \cite{potasscoManual}.
Although we provide essential definitions of CSPs in the next section, 
we refer the reader to the literature 
\cite{dechter03,CPHandbook} for a broader perspective.


\section{Background}\label{sec:background}

A \emph{Constraint Satisfaction Problem} (CSP)
is given by a pair $
(\Var,\Con)$ consisting of
a set~$\Var$ of \emph{variables} and 
a set~$\Con$ of \emph{constraint clauses}.
Every variable $x\in\Var$ has an associated finite \emph{domain} $\dom{x}$ such that
either $\dom{x}=\{\top,\bot\}$ or $\emptyset\subset\dom{x}\subseteq\mathbb{Z}$;
$x$ is a \emph{Boolean variable} if $\dom{x}=\{\top,\bot\}$,
and an \emph{Integer variable} otherwise.
We denote the set of Boolean variables in~$\Var$ by $\Bool$ and the 
set of Integer variables in~$\Var$ by $\Int$.
A constraint clause $C\in\Con$ is a set of literals over
Boolean variables in $\Bool$ as well as
linear inequalities or global constraints on Integer variables in $\Int$.
Any \emph{literal} in~$C$ is of the form $e$ or $\opp{e}$,
where $e$ is either a Boolean variable in $\Bool$,
a linear inequality, or a global constraint.
A \emph{linear inequality} is an expression
$\sum_{1\leq i\leq n}a_ix_i\leq m$ in which
$m$ as well as all $a_i$ for $1\leq i\leq n$ are Integer constants
and $x_1,\dots,x_n$ are Integer variables in $\Int$.
A \emph{global constraint} (cf.\ \cite{belsim11a})
is an arbitrary relation over Integer variables in $\Int$;
we here restrict ourselves to table and alldifferent constraints
over subsets $\{x_1,\dots,x_n\}$ of the Integer variables in $\Int$,
where a  \emph{table constraint} specifies tuples
$(d_1,\dots,d_n)\in \dom{x_1}\times\dots\times\dom{x_n}$ of admitted value
combinations
and \emph{alldifferent} applies if $x_1,\dots,x_n$ are assigned to
distinct values in their respective domains.\footnote{%
Linear inequalities relying on further comparison operators,
such as $<$, $>$, $\geq$, $=$, and $\neq$,
can be converted into the considered format via appropriate replacements
\cite{tatakiba09a}.
Moreover, note that we here limit the consideration of global constraints
to the ones that are directly, i.e., without normalization by \sugar,
supported in our prototypical ASP encodings shipped with \aspartame.}

Given a CSP $(\Var,\Con)$, a \emph{variable assignment}~$\ass$
is a (total) mapping $\ass:\Var\rightarrow\bigcup_{x\in\Var}\dom{x}$
such that $\val{x}\in\dom{x}$ for every $x\in\Var$.
A Boolean variable $x\in\Bool$ is \emph{satisfied} w.r.t.~$\ass$ if $\val{x}=\top$.
Likewise, a linear inequality $\sum_{1\leq i\leq n}a_ix_i\leq m$ is
\emph{satisfied} w.r.t.~$\ass$
if $\sum_{1\leq i\leq n}a_i\val{x_i}\leq m$. 
Table constraints $e\subseteq\dom{x_1}\times\dots\times\dom{x_n}$ and
alldifferent constraints over subsets $\{x_1,\dots,x_n\}$ of $\Int$
are \emph{satisfied} w.r.t.~$\ass$ if $(\val{x_1},\dots,\val{x_n})\in e$ or
$\val{x_i}\neq\val{x_j}$ for all $1\leq i<j\leq n$, respectively.
Any Boolean variable, linear inequality, or global constraint that is not
satisfied w.r.t.~$\ass$ is \emph{unsatisfied} w.r.t.~$\ass$.
A constraint clause $C\in\Con$
is \emph{satisfied} w.r.t.~$\ass$ if there is some literal $e\in C$
(or $\opp{e}\in C$) such that $e$ is satisfied (or unsatisfied) w.r.t.~$\ass$.
The assignment $\ass$ is a \emph{solution} for $(\Var,\Con)$ if
every $C\in\Con$ is satisfied w.r.t.~$\ass$.

\begin{example}\label{ex}
Consider a CSP $
(\Var,\Con)$ with Boolean and Integer variables
$\Bool=\{b\}$ and $\Int=\{x,y,z\}$,
where $\dom{x}=\dom{y}=\dom{z}=\{1,2,3\}$,
and 
constraint clauses $\Con=\{C_1,C_2,C_3\}$ 
as follows:
\begin{align}
\label{ex:alldifferent}
C_1 & {} =
\left\{\diff(x,y,z)\right\}
\\
\label{ex:inequality}
C_2 & {} =
\left\{b,4x-3y+z\leq 0\right\}
\\
\label{ex:table}
C_3 & {} =
\left\{\opp{b},(x,y)\in\{(1,3),(2,2),(3,1)\}\right\}
\end{align}
The alldifferent constraint in $C_1$ requires values assigned to
$x$, $y$, and $z$ to be mutually distinct.
Respective assignments~$\ass$ 
satisfying the linear inequality $4x-3y+z\leq 0$ in $C_2$ include 
$\val{x}=2$, $\val{y}=3$, and $\val{z}=1$ or
$\val{x}=1$, $\val{y}=3$, and $\val{z}=2$,
while the table constraint in~$C_3$ is satisfied w.r.t.\
assignments~$\ass$ containing
$\val{x}=1$, $\val{y}=3$, and $\val{z}=2$ or
$\val{x}=3$, $\val{y}=1$, and $\val{z}=2$.
In view of the Boolean variable~$b$,
whose value allows for ``switching'' between the linear inequality 
in $C_2$ and the table constraint in $C_3$, we 
obtain the following 
solutions $\ass_1,\dots,\ass_4$ for $(\Var,\Con)$:

\begin{tabular}{|@{\,}r@{\,}||@{\,}c@{\,}|@{\;}c@{\;}|@{\;}c@{\;}|@{\;}c@{\;}|}
\cline{1-5}
 & $b$ & $x$ & $y$ & $z$
\\\hline\hline
$\ass_1$ & $\bot$ & $2$ & $3$ & $1$
\\\cline{1-5}
$\ass_2$ & $\bot$ & $1$ & $3$ & $2$
\\\cline{1-5}
$\ass_3$ & $\top$ & $1$ & $3$ & $2$
\\\cline{1-5}
$\ass_4$ & $\top$ & $3$ & $1$ & $2$
\\\cline{1-5}
\end{tabular}
\end{example}



\section{Approach}\label{sec:approach}

The \aspartame\ tool extends the SAT-based solver \sugar\ by
an output component to represent a CSP in terms of ASP facts.
The generated facts can then, as usual, be combined with a
first-order encoding processable with off-the-shelf ASP systems.
In what follows, we describe the format of facts generated
by \aspartame, and we present a dedicated 
ASP encoding
utilizing function terms to capture substructures in CSP instances.

\subsection{Fact Format}\label{sec:fact}

Facts express the variables and constraints of a CSP instance in the syntax of
ASP grounders like \gringo\ \cite{potasscoManual}.
Their format is easiest explained on the CSP 
from Example~\ref{ex}, whose fact representation is shown in Listing~\ref{ex:facts}.
While facts of the predicate \code{var}/2
provide labels of Boolean variables like~$b$,
the predicate \code{var}/3 includes a third argument for declaring
the domains of Integer variables like $x$, $y$, and $z$.
Domain declarations rely on function terms \code{range(}$l$\code{,}$u$\code{)},
standing for continuous Integer intervals~$[l,u]$.
While one term, \code{range(1,3)}, suffices for the common domain $\{1,2,3\}$
of $x$, $y$, and $z$, in general,
several intervals can be specified (via separate facts)
to form non-continuous domains.
Note that the interval format for Integer domains offers a compact
fact representation of 
(continuous) domains; e.g.,
the single term \code{range(1,10000)} captures a domain with $10000$ elements.
Furthermore, the usage of meaningful function terms
avoids any need for artificial labels to refer to domains or parts thereof.%
\begin{lstlisting}[float=t,caption={Facts representing the CSP % $\CSP$
from Example~\ref{ex}.},captionpos=b,frame=single,label=ex:facts,%
basicstyle=\ttfamily\scriptsize]
var(bool,b).  var(int,x;y;z,range(1,3)).

constraint(1,global(alldifferent,arg(x,arg(y,arg(z,nil))))).
constraint(2,b).
constraint(2,op(le,op(add,op(add,op(mul,4,x),op(mul,-3,y)),op(mul,1,z)),0)).
constraint(3,op(neg,b)).
constraint(3,rel(r,arg(x,arg(y,nil)))).

rel(r,2,3,supports).
tuple(r,1,1,1).  tuple(r,1,2,3).
tuple(r,2,1,2).  tuple(r,2,2,2).
tuple(r,3,1,3).  tuple(r,3,2,1).
\end{lstlisting}

The literals of constraint clauses 
are also represented by means of function terms.
In fact, the second argument of \code{constraint}/2 in
Line~3 of Listing~\ref{ex:facts} stands for 
$\diff(x,y,z)$ from the constraint clause~$C_1$ in~(\ref{ex:alldifferent}),
which is identified via the first argument of \code{constraint}/2.
Since every fact of the predicate \code{constraint}/2 is supposed to describe
a single literal only, constraint clause identifiers establish the
connection between individual literals of a clause.
This can be observed on the facts in Line~4--7, specifying literals belonging to
the binary constraint clauses~$C_2$ and~$C_3$ in~(\ref{ex:inequality}) and~(\ref{ex:table}).
Here, the terms \code{b} and \code{op(neg,b)} refer to the literals~$b$ and~$\opp{b}$
over Boolean variable~$b$, where \code{op(neg,}$e$\code{)} is the general notation
of~$\opp{e}$ for all (supported) constraint expressions~$e$.
The more complex term of the form \code{op(le,}$\Sigma$\code{,}$m$\code{)}
in Line~5 stands for a linear inequality $\Sigma\leq m$.
In particular,
the inequality $4x-3y+z\leq 0$ from~$C_2$ is represented by
nested \code{op(add,}$\Sigma$\code{,}$ax$\code{)} terms whose last argument $ax$
and deepest~$\Sigma$ part are of the form \code{op(mul,}$a$\code{,}$x$\code{)};
such nesting corresponds to the precedence $(((4*x)+(-3*y))+(1*z))\leq 0$.
The representation by function terms captures linear inequalities of arbitrary arity and,
as with Integer intervals, associates (sub)sums with canonical labels.
Currently, the order of arguments~$ax$ is by variable labels~$x$,
while more ``clever'' orders may be established in the future.

The function terms expressing table and alldifferent constraints both
include an argument list of the form
\code{arg(}$x_1$\code{,arg(}$\dots$\code{,arg(}$x_n$\code{,nil)}$\dots$\code{))},
in which $x_1,\dots,x_n$ refer to Integer variables.
In Line~3 of Listing~\ref{ex:facts}, an alldifferent constraint over
arguments~$\vec{x}$ is declared via \code{global(alldifferent,}$\vec{x}$\code{)};
at present, \code{alldifferent} is a fixed keyword in facts generated by \aspartame,
but support for other kinds of global constraints can be added in the future.
Beyond an argument list~$\vec{x}$,
function terms of the form \code{rel(}$r$\code{,}$\vec{x}$\code{)} also
include an identifier~$r$ referring to a collection of
table constraint tuples.
For instance, the corresponding argument~\code{r} in Line~7 addresses the
tuples specified by the facts in Line~9--12.
Here, \code{rel(r,2,3,supports)} declares that \code{r} is of arity~$2$ and
includes $3$ tuples, provided as white list entries via facts of the form
\code{tuple(r,}$t$\code{,}$i$\code{,}$d$\code{)}.
The latter include tuple and argument identifiers~$t$ and~$i$ along with
a value~$d$.
Accordingly,
the facts in Line~10,~11, and~12 specify the pairs $(1,3)$, $(2,2)$, and $(3,1)$
of values, which are the combinations admitted by
the table constraint from~$C_3$ in~(\ref{ex:table}).
The application of the table constraint to variables~$x$ and~$y$ is expressed by
the argument list in Line~7, so that tuple declarations can be re-used
for other variables subject to a similar table constraint.

\subsection{First-Order Encoding}\label{sec:encoding}

In addition to an  output component extending \sugar\ for generating ASP facts,
\aspartame\ comes along with alternative first-order ASP encodings of
solutions for CSP instances.
In the following, we sketch a dedicated encoding that, for one, relies on
function terms to capture recurrences of similar structures and,
for another, lifts the order encoding approach to structural subsum entities.

\subsubsection{Static Extraction of Relevant Values}

To begin with,
Listing~\ref{ex:domain} shows (relevant) instances of domain predicates,
evaluated upon grounding, for the CSP 
from Example~\ref{ex}.
While derived facts in Line~1 merely provide a projection of the
predicate \code{var}/3 omitting associated domains, the instances of
\code{look}/2 in Line~2 express that all values in the common
domain $\{1,2,3\}$ of $x$, $y$, and $z$ shall be considered.
In fact, domain predicates extract variable values that can be
relevant for the satisfiability of a CSP instance, while discarding the rest.
The respective static analysis consists of three stages:
(i) isolation of threshold values relevant to linear inequalities;
(ii) addition of missing values for variables occurring in 
      alldifferent
      constraints;
(iii)
addition 
of white/black list values for table constraints.

In the first stage,
we consider the domains of Integer variables~$x$
in terms of corresponding (non-overlapping) intervals
$\range{x}=\{[l_1,u_1],\dots,[l_k,u_k]\}$.
These are extended to multiplications by Integer constants~$a$
according to the following scheme:
\begin{equation*}
\range{ax} =
\left\{
\begin{array}{l@{\quad\text{if \ }}l}
\{[a*l_1,a*u_1],\dots,[a*l_k,a*u_k]\} &
0\leq a
\\
\{[a*u_k,a*l_k],\dots,[a*u_1,a*l_1]\} &
a < 0
\end{array}
\right.
\end{equation*}
For 
$4x-3y+z\leq 0$ from~$C_2$ in~(\ref{ex:inequality}),
we 
get $\range{4x}=\{[4,12]\}$, $\range{-3y}=\{[-9,-3]\}$, and $\range{1z}=\{[1,3]\}$.
Such intervals are used to retrieve bounds for (sub)sums:
\begin{align*}
\low{ax}={}& \min{\{l\mid [l,u]\in\range{ax}\}}
\\
\upp{ax}={}& \max{\{u\mid [l,u]\in\range{ax}\}}
\\
\low{a_1x_1+a_2x_2}={}& \low{a_1x_1}+\low{a_2x_2}
\\
\upp{a_1x_1+a_2x_2}={}& \upp{a_1x_1}+\upp{a_2x_2}
\end{align*}
Given $\low{4x}=4$, $\upp{4x}=12$, $\low{-3y}=-9$, $\upp{-3y}=-3$, $\low{1z}=1$,
and $\upp{1z}=3$,
we derive 
$\low{4x-3y}=-5$,
$\upp{4x-3y}=9$,
$\low{4x-3y+z}=-4$, and
$\upp{4x-3y+z}=12$.

In view of the comparison with~$0$ in $4x-3y+z\leq 0$,
we can now ``push in'' relevant thresholds via:
\pagebreak[1]%
\begin{align*}
\blow{\mbox{$\sum$}_{1\leq i\leq n}a_ix_i}&{}=\max\{m,\low{\mbox{$\sum$}_{1\leq i\leq n}a_ix_i}\}
\phantom{\min\upp{}}\!\!\! \text{for \ }\mbox{$\sum$}_{1\leq i\leq n}a_ix_i\leq m
\\
\bupp{\mbox{$\sum$}_{1\leq i\leq n}a_ix_i}&{}=\min\{m,\upp{\mbox{$\sum$}_{1\leq i\leq n}a_ix_i}\}
\phantom{\max\low{}}\!\!\! \text{for \ }\mbox{$\sum$}_{1\leq i\leq n}a_ix_i\leq m
\\
\blow{\mbox{$\sum$}_{1\leq i\leq n-1}a_ix_i}&{}= \max\{\blow{\mbox{$\sum$}_{1\leq i\leq n}a_ix_i}-\upp{a_nx_n},\low{\mbox{$\sum$}_{1\leq i\leq n-1}a_ix_i}\}
\\
\bupp{\mbox{$\sum$}_{1\leq i\leq n-1}a_ix_i}&{}= \min\{\bupp{\mbox{$\sum$}_{1\leq i\leq n}a_ix_i}-\low{a_nx_n},\upp{\mbox{$\sum$}_{1\leq i\leq n-1}a_ix_i}\}
\end{align*}
Such threshold analysis leads to
$\blow{4x-3y+z}=\bupp{4x-3y+z}=0$,
$\blow{4x-3y}=-3$,
$\bupp{4x-3y}=-1$,
$\blow{4x}=4$, and
$\bupp{4x}=8$,
telling us that subsums relevant
for checking whether $4x-3y+z\leq 0$
satisfy
$-3\leq 4x-3y\leq -1$
and
$4\leq 4x\leq 8$.
Note that maxima (or minima) used to construct $\blow{\sum_{1\leq i\leq n}a_ix_i}$ 
(or $\bupp{\sum_{1\leq i\leq n}a_ix_i}$)
serve two purposes.
For one, they
correct infeasible arithmetical thresholds to domain values; e.g.,
$\blow{4x-3y}-\upp{-3y}=-3+3=0$ tells us that $0$ would be the greatest lower bound to consider
for $4x$
(since $4x-3y+z\leq 0$ were necessarily satisfied when $4x\leq 0$),
while the smallest possible value $\low{4x}=4$ exceeds~$0$. 
For another, dominating values like $4x=12$ are discarded, given that
$4x-3y+z\leq 0$ cannot hold when
$4x>\bupp{4x-3y}-\low{-3y}=-1+9=8$.

Letting $\bound{0}=\{0\}$, 
the upper bounds
for $\sum_{1\leq i\leq n}a_ix_i$
that
deserve further consideration are then obtained as follows:
\begin{align*}
& & \SQUEEZE
\bound{\mbox{$\sum$}_{1\leq i\leq n}a_ix_i} =
\{
  \max\{j+a_n{*}k,\blow{\mbox{$\sum$}_{1\leq i\leq n}a_ix_i}\}
  \mid
j\in\bound{\mbox{$\sum$}_{1\leq i\leq n-1}a_ix_i},
{} \\ & & \SQUEEZE
k\in\mathbb{Z},
  [l,u]\in\range{a_nx_n},
  l\leq a_n{*}k \leq \min\{u,\bupp{\mbox{$\sum$}_{1\leq i\leq n}a_ix_i}-j\}
\}
\end{align*}
Starting from the above thresholds,
$\bound{4x}=\{4,8\}$,
$\bound{4x-3y}=\{-3,-2,-1\}$, and
$\bound{4x-3y+z}=\{0\}$
indicate upper bounds for subsums
that are of interest
in evaluating $4x-3y+z\leq 0$.
Upper bounds in $\bound{\mbox{$\sum$}_{1\leq i\leq n}a_ix_i}$
can in turn be related to ``maximal'' pairs of addends:
\begin{align*}
& & \SQUEEZE
\add{\mbox{$\sum$}_{1\leq i\leq n}a_ix_i}
=
\{
(
  j,
  \max\{
a_n{*}k \mid 
[l,u]\in\range{a_nx_n},l\leq a_n{*}k\leq\min\{u,\ub-j\}
,{}
\\ & & \SQUEEZE
k\in\mathbb{Z}
\}) \mid
j\in\bound{\mbox{$\sum$}_{1\leq i\leq n-1}a_ix_i},
\ub\in\bound{\mbox{$\sum$}_{1\leq i\leq n}a_ix_i},
\low{a_nx_n}\leq\ub-j
\}
\end{align*}
In our example, we get
$\add{4x}=\{(0,4),(0,8)\}$,
$\add{4x-3y}=\{(4,-9),(4,-6),(8,-9)\}$, and
$\add{4x-3y+z}=\{(-3,3),(-2,2),(-1,1)\}$.

Finally, we associate each pair
$(j,a_n{*}k)\in\add{\mbox{$\sum$}_{1\leq i\leq n}a_ix_i}$
of addends
with the
upper bound
$\erg{j,a_n{*}k}=
 \min\{\ub\in\bound{\mbox{$\sum$}_{1\leq i\leq n}a_ix_i} \mid
       j+a_n{*}k\leq\ub\}$,
thus obtaining
$\erg{0,4}=4$, $\erg{0,8}=8$,
$\erg{4,-9}=-3$, $\erg{4,-6}=-2$, $\erg{8,-9}=-1$,
and $\erg{-3,3}=\erg{-2,2}=\erg{-1,1}=0$.
\begin{lstlisting}[float=t,caption={Domain predicates derived via stratified rules
(not shown) from % the 
facts in Listing~\ref{ex:facts}.},captionpos=b,frame=single,label=ex:domain,%
basicstyle=\ttfamily\scriptsize]
var(int,x;y;z).
look(x;y;z,1;2;3).

order(x;y;z,3,2).
order(x;y;z,2,1).

order(op(add,op(mul,4,x),op(mul,-3,y)),-1,-2).
order(op(add,op(mul,4,x),op(mul,-3,y)),-2,-3).

look(op(mul,4,x),1;2,1).
look(op(mul,-3,y),2;3,-1).
look(op(mul,1,z),1;2;3,1).

look(op(add,op(mul,4,x),op(mul,-3,y)),4,-6,-2).
look(op(add,op(mul,4,x),op(mul,-3,y)),4,-9,-3).
look(op(add,op(mul,4,x),op(mul,-3,y)),8,-9,-1).

look(op(add,op(add,op(mul,4,x),op(mul,-3,y)),op(mul,1,z)),-1,1,0).
look(op(add,op(add,op(mul,4,x),op(mul,-3,y)),op(mul,1,z)),-2,2,0).
look(op(add,op(add,op(mul,4,x),op(mul,-3,y)),op(mul,1,z)),-3,3,0).

bound(op(add,op(add,op(mul,4,x),op(mul,-3,y)),op(mul,1,z)),12).

difind(arg(x,arg(y,arg(z,nil))),x,1).
difind(arg(x,arg(y,arg(z,nil))),y,2).
difind(arg(x,arg(y,arg(z,nil))),z,3).
difmax(arg(x,arg(y,arg(z,nil))),3,1;2;3).
difall(arg(x,arg(y,arg(z,nil)))).

relind(r,arg(x,arg(y,nil)),x,1).
relind(r,arg(x,arg(y,nil)),y,2).
\end{lstlisting}

The described analysis of thresholds for subsums
is implemented via deterministic domain predicates in our ASP encoding.
Variables' domain values underlying relevant addends are provided
by the derived facts in Line~10--12 of Listing~\ref{ex:domain}.
Note that value~$3$ for~$x$ as well as~$1$ for~$y$
are ignored here,
given that $4x=12$ and $-3y=-3$ do not admit
$4x-3y+z\leq 0$ to hold.
The mapping of relevant addends to their associated upper bound
can be observed in Line~14--20 for the (sub)sums $4x-3y$ and
$(4x-3y)+z$.
The respective facts describe patterns for mapping assigned domain
values to their multiplication results and then to upper bounds for
subsums, which are eventually subject to a (non-trivial)
comparison in some linear inequality.
(Trivial comparisons are performed via the total upper bound
 for an addition result, as given in Line~22.)
Notably, the static threshold analysis is implemented on terms
representing the domains of variables, and outcomes are then mapped
back to original variables.
Thus, linear inequalities over different
variables with the same domains are analyzed only once.
The final function terms, however, mention the variables whose values
are evaluated, where recurring substructures may share a common term
with which all relevant threshold values are associated.

Although the analysis of the linear inequality $4x-3y+z\leq 0$ identifies
the values~$3$ for~$x$ and~$1$ for~$y$ as redundant,
the presence of $\diff(x,y,z)$
leads to their ``release'' as relevant candidates for~$x$ and~$y$.
Accordingly, all values in the common
domain $\{1,2,3\}$ of $x$, $y$, and $z$
are put into (decreasing) order, given by the derived facts in Line~4--5.
Beyond that, the order among
relevant upper bounds in $\bound{4x-3y}=\{-3,-2,-1\}$
is reflected in Line~7--8;
this is used to apply the order encoding to
structural subsum variables (in addition to the input variables~$x$, $y$, and~$z$).
The residual derived facts in Line~24--31 serve convenience by 
associating indexes to the arguments of $\diff(x,y,z)$
as well as to~$x$ and~$y$ considered in $(x,y)\in\{(1,3),(2,2),(3,1)\}$.
Furthermore,
the fact in Line~27 indicates the index~$3$ of variable~$z$
in $\diff(x,y,z)$ as the final position at which either of the values~$1$, $2$, or~$3$
can possibly be assigned,
and the fact in Line~28 expresses that all three values 
in $\dom{x}\cup\dom{y}\cup\dom{z}=\{1,2,3\}$ must be assigned
in order to satisfy $\diff(x,y,z)$.

\subsubsection{Non-deterministic Encoding Part}
\begin{lstlisting}[float=t,caption={First-order encoding of solutions
for finite linear CSPs.},captionpos=b,frame=single,label=ex:encode,%
basicstyle=\ttfamily\scriptsize]
% generate variable assignment

{ less(V,E) : order(V,E,_) } :- var(int,V).
 :- order(V,E1,E2), less(V,E2), not less(V,E1).
value(V,E) :- look(V,E), not less(V,E), less(V,EE) : order(V,EE,E).

{ value(V,true) } :- var(bool,V).
value(V,false) :- var(bool,V), not value(V,true).

% evaluate linear inequalities

leq(op(mul,F,V),F*E) :- look(op(mul,F,V),E,1), less(V,EE) : order(V,EE,E).
leq(op(mul,F,V),F*E) :- look(op(mul,F,V),E,-1), not less(V,E).
leq(op(add,S1,S2),E) :- look(op(add,S1,S2),E1,E2,E), leq(S1,E1;;S2,E2).
leq(op(add,S1,S2),E) :- order(op(add,S1,S2),E,EE), leq(op(add,S1,S2),EE).

% evaluate alldifferent expressions

seen(A,I,E) :- difind(A,V,I), value(V,E).
seen(A,I,E) :- difind(A,_,I), seen(A,I-1,E), not difmax(A,I-1,E).

redo(A) :- difind(A,V,I), seen(A,I-1,E), value(V,E).
redo(A) :- difall(A), difmax(A,I,E), not seen(A,I,E).

% evaluate table expressions

rela(R,A,T,2)   :- relind(R,A,V,1), tuple(R,T,1,E), value(V,E).
rela(R,A,T,I+1) :- relind(R,A,V,I), tuple(R,T,I,E), value(V,E), rela(R,A,T,I).

rela(R,A,U) :- rela(R,A,_,I+1), rel(R,I,_,U).

% check constraint clauses

hold(C) :- constraint(C,V), value(V,true).
hold(C) :- constraint(C,op(neg,V)), value(V,false).

hold(C) :- constraint(C,op(le,S,E)), bound(S,U), leq(S,E) : E < U.
hold(C) :- constraint(C,op(neg,op(le,S,E))), bound(S,U), E < U, not leq(S,E).

hold(C) :- constraint(C,global(alldifferent,A)), not redo(A).
hold(C) :- constraint(C,op(neg,global(alldifferent,A))), redo(A).

hold(C) :- constraint(C,rel(R,A)), not rela(R,A,conflicts),
           rela(R,A,supports) : rel(R,_,_,supports).
hold(C) :- constraint(C,op(neg,rel(R,A))), not rela(R,A,supports),
           rela(R,A,conflicts) : rel(R,_,_,conflicts).

constraint(C) :- constraint(C,_).
 :- constraint(C), not hold(C).

% display variable assignment

#hide.
#show value/2.
\end{lstlisting}
With the described domain predicates at hand,
the encoding part in Listing~\ref{ex:encode}
implements the non-deterministic guessing of a variable assignment along with
the evaluation of constraint clauses.
Following the idea of order encodings in SAT \cite{crabak94a,tatakiba09a},
the choice rule in Line~3 permits guessing \code{less(V,E)}
for all but the smallest (relevant) value~\code{E}
in the domain of an Integer variable~\code{V},
thus indicating that~\code{V} is assigned to some
smaller value than~\code{E}.
The consistency among guessed atoms is established by the
integrity constraint in Line~4,
requiring \code{less(V,E1)} 
to hold if \code{less(V,E2)} is true
for the (immediate) predecessor 
value~\code{E2}  of~\code{E1}.
The actual value assigned to~\code{V},
given by the greatest~\code{E} for which \code{less(V,E)} is false,
is extracted in Line~5.
For Boolean variables,
the value~\code{true}
can be guessed unconditionally via 
the choice rule in Line~7, and
\code{false} is derived otherwise via the rule in Line~8.

The dedicated extension of the order encoding idea to 
subsums of linear inequalities is implemented by means of the
rules in Line~12--15 of Listing~\ref{ex:encode}.
To this end, upper bounds for
singular multiplication results indicated as
relevant by instances of \code{look(op(mul,F,V),E,G)}
are directly derived from \code{less}/2.
Thereby, the flag \code{G}${}={}$\code{F}${}/{}$\code{|F|}
provides the polarity of the actual coefficient~\code{F}.\footnote{%
Coefficients given in facts generated by \aspartame\ are distinct from~$0$.}
If~\code{F} is positive, i.e., \code{G}${}={}$\code{1}, the upper bound 
\code{F*E} is established as soon as 
\code{less(V,EE)} holds for the immediate successor value~\code{EE}  of~\code{E}
(or if \code{E} is the greatest relevant value in the domain of~\code{V}).
On the other hand, if \code{G}${}={}$\code{-1} indicates that~\code{F} is negative,
the upper bound~\code{F*E} is derived from \code{not}~\code{less(V,E)},
which means that the value assigned to~\code{V} is greater than or equal to~\code{E}.
Relevant upper bounds~\code{E} for subsums
rely on maximal pairs~$($\code{E1}$,$\code{E2}$)$
of addends, identified via static threshold analysis and readily
provided by instances of \code{look(op(add,S1,S2),E1,E2,E)}.
In fact,
the rule in Line~14 derives \code{leq(op(add,S1,S2),E)},
indicating that \code{S1}${}+{}$\code{S2}${}\leq{}$\code{E},
from \code{leq(S1,E1)} and \code{leq(S2,E2)}.
Although an established upper bound inherently implies
any greater (relevant) upper bound to hold as well
w.r.t.\ a total variable assignment,
ASP (and SAT) solvers are not committed to guessing
``input variables'' first.
Rather, structural variables like the instances of \code{leq(op(add,S1,S2),E)}
may be fixed upon solving, possibly in view of recorded conflict clauses,
before a total assignment has been determined.
In view of this, the additional rule in Line~15 makes sure 
that an established upper bound~\code{EE} propagates to its
immediate successor~\code{E} (if there is any).
For instance, (simplified) ground instances of the rule
stemming from $\bound{4x-3y}=\{-3,-2,-1\}$ include the following:

\begin{tabular}{l}
\mbox{\code{leq(op(add,op(mul,4,x),op(mul,-3,y)),-1) :-}}
\\\qquad\mbox{\code{leq(op(add,op(mul,4,x),op(mul,-3,y)),-2).}}
\\\mbox{\code{leq(op(add,op(mul,4,x),op(mul,-3,y)),-2) :-}}
\\\qquad\mbox{\code{leq(op(add,op(mul,4,x),op(mul,-3,y)),-3).}}
\end{tabular}

\noindent
Unlike with the domains of Integer variables,
we rely on a rule, rather than an integrity constraint,
to establish consistency among the bounds for structural subsums.
The reason for this is that upper bounds for
addends \code{S1} and \code{S2},
contributing left and right justifications, 
may include divergent gaps,
so that consistent value orderings for them
are, in general, not guaranteed to immediately
produce all relevant upper bounds for \code{S1}${}+{}$\code{S2}.
Encoding variants resolving this issue and
using integrity constraints like the one in Line~4
are a subject to future investigation.

While linear inequalities can be evaluated by means of boundaries
derived more or less directly from instances of \code{less(V,E)},
the evaluation of alldifferent and table constraints in Line~19--23
and Line~27--30 of Listing~\ref{ex:encode} relies on
particular instances of \code{value(V,E)}.
The basic idea of checking whether an alldifferent constraint holds
is to propagate assigned values along the indexes of participating variables.
Then, a recurrence is detected when the value assigned to a
variable with index~\code{I} has been marked as already assigned,
as determined from \code{seen(A,I-1,E)} in Line~22.
Moreover, whenever \code{difall(A)} indicates that all domain values
for the variables in argument list~\code{A} must be assigned,
the rule in Line~23 additionally derives a recurrence from some gap
(a value that has not been assigned to the variable at the last possible index).
Our full encoding further features so-called ``pigeon-hole constraints''
(cf.\ \cite{drewal10a,mecost13a}) to check that the smallest or greatest
$1,\dots,n-1$ domain values for an alldifferent constraint with~$n$ variables
are not populated by more than $i$ variables for $1\leq i \leq n-1$.
Such conditions can again be checked based on instances of \code{less(V,E)},
and both counter-based (cf.\ \cite{conf/cp/Sinz05}) as well as
aggregate-based (cf.\ \cite{siniso02a}) implementations are applicable
in view of the native support of aggregates by ASP solvers like \clasp\
(cf.\ \cite{gekakasc09a}).
In fact, the usage of rules to express redundant constraints,
like the one in Line~23 or those for pigeon-hole constraints,
as well as their ASP formulation  provide various degrees of freedom,
where 
comprehensive evaluation and 
configuration methods
are subjects to future work.

The strategy for evaluating table constraints is closely related to
the one for detecting value recurrences in  alldifferent constraints.
Based on the indexes of variables in a table constraint,
tuples that are (still) admissible are forwarded via the rules in
Line~27--28.
The inclusion of a full tuple in an assignment is
detected by the rule in Line~30,
checking whether the arity~\code{I} of a table constraint
has been reached for some tuple,
where a   value \code{supports} or \code{conflicts} for~\code{U}
additionally indicates whether the included tuple belongs to
a white or black list, respectively.
Note that this strategy avoids explicit references to 
variables whose values are responsible for the exclusion of tuples,
given that lack of inclusion is detected from incomplete tuple traversals.

Finally, the rules in Line~34--49 explore the values 
assigned to Boolean variables
and the outcomes of evaluating particular kinds of constraints
to derive \code{hold(C)} if and only if
some positive or negative literal in~\code{C} is satisfied or
unsatisfied, respectively,
w.r.t.\ the variable assignment represented by instances 
of \code{value(V,E)}.
Without going into details, let us still note that our full encoding
also features linear inequalities relying on the comparison operators
$\geq$, $=$, and~$\neq$, 
for which additional rules are included to derive \code{hold(C)},
yet sticking to the principle of upper bound evaluation via \code{leq}/2.
In fact, the general possibility of 
complemented constraint expressions as well as of disjunctions
potentially admits unsatisfied constraint
expressions w.r.t.\ solutions, and our encoding reflects this
by separating the evaluation of particular constraint expressions in Line~12--30
from further literal and clause evaluation in Line~34--49.


\section{The \aspartame\ System}\label{sec:system}

The architecture of the \aspartame\
system is given in Figure~\ref{fig:arch}.
As mentioned,
\aspartame\ re-uses \sugar's front-end for parsing and normalizing CSPs.
Hence, it accepts the same input formats, 
viz.\ 
XCSP\footnote{\url{\texttt{http://www.cril.univ-artois.fr/CPAI08/XCSP2\_1.pdf}}}
and \sugar's native CSP format\footnote{\url{\texttt{http://bach.istc.kobe-u.ac.jp/sugar/package/current/docs/syntax.html}}}.
We then implemented an output hook for \sugar\ that provides us with the resulting CSP instance in
the fact format described in Section~\ref{sec:fact}. 
These facts are then used for grounding the (full version of the) dedicated
ASP encoding in Listing~\ref{ex:encode} or an alternative one (discussed below).
This is done by the ASP grounder \gringo.
In turn, the resulting propositional logic program is passed to the ASP solver \clasp\ that returns
an assignment, representing a solution to the original CSP instance.

\begin{figure}[t]
  \thicklines
  \setlength{\unitlength}{1.28pt}
  \small
  \begin{picture}(280,57)(4,-10)
    \put( 6, 20){\dashbox(24,24){\shortstack{CSP\\Instance}}}
    \put( 40, 15){\framebox(50,30){\sugar\quad}}
    \put( 80, 15){\dashbox(10,30){\it\shortstack{A\\S\\P}}}
    \put(100, 20){\dashbox(20,20){\shortstack{ASP\\Facts}}}
    \put( 80,-15){\dashbox(40,20){\shortstack{ASP\\Encoding}}}
    \put(130, 15){\framebox(50,30){\gringo}}
    \put(190, 15){\framebox(50,30){\clasp}}
    \put(250, 20){\dashbox(24,20){\shortstack{CSP\\Solution}}}
    \put( 30, 30){\vector(1,0){10}}
    \put( 90, 30){\vector(1,0){10}}
    \put(120, 30){\vector(1,0){10}}
    \put(180, 30){\vector(1,0){10}}
    \put(240, 30){\vector(1,0){10}}
    \put(120, -5){\line(1,0){4}}
    \put(124, -5){\line(0,1){35}}
  \end{picture}  
\caption{Architecture of \aspartame.}
\label{fig:arch}
\end{figure}

We 
empirically access the performance of \aspartame\
relative to two ASP encodings, 
the dedicated one described in Section~\ref{sec:encoding} as well as
a more direct encoding inspired by the original CNF construction
of \sugar\ \cite{tatakiba09a},
and additionally consider
the SAT-based reference solver \sugar\ (2.0.0).
In either case, we use the combined ASP and SAT solver \clasp\ (2.1.0),
and ASP-based approaches further rely on \gringo\ (3.0.5)
for grounding ASP encodings on facts generated by \aspartame.
We selected 60 representative CSP instances (that are neither too easy nor too hard),
consisting of intensional and global constraints,
from the benchmarks of the 
2009 CSP Competition\footnote{%
\url{\texttt{http://www.cril.univ-artois.fr/CPAI09}}}
for running systematic experiments on a cluster of Linux machines equipped with
dual Xeon E5520 quad-core 2.26 GHz processors and 48 GB RAM.
To get some first insights into suitable search options,
we ran \clasp\ with its default (berkmin-like) and the popular ``vsids''
decision heuristic;
while SAT-based preprocessing (cf.\ \cite{eenbie05a}) 
is performed by default on CNF inputs,
we optionally enabled it for (ground) ASP instances,
leading to four combinations of \clasp\ settings for ASP-based approaches
and two for SAT-based \sugar.
\begin{table}
  \caption{Experiments comparing ASP encoding variants and 
           the SAT-based solver \sugar.\label{tab:experiments}}
\centering%
{\scriptsize
\newcommand{\HACK}{\hline}
\begin{tabular}{|@{\,}l@{\,}||@{\,}r@{\,}||@{\,}r@{\,}|@{\,}r@{\,}|@{\,}r@{\,}|@{\,}r@{\,}|@{\,}r@{\,}||@{\,}r@{\,}|@{\,}r@{\,}|@{\,}r@{\,}|@{\,}r@{\,}|@{\,}r@{\,}||@{\,}r@{\,}|@{\,}r@{\,}|@{\,}r@{\,}|}
	\hline
        &
	&\multicolumn{5}{@{\,}c@{\,}||@{\,}}{\textit{ASP Encoding 1 (dedicated)}}
	&\multicolumn{5}{@{\,}c@{\,}||@{\,}}{\textit{ASP Encoding 2 (SAT-inspired)}}
	&\multicolumn{3}{@{\,}c@{\,}|}{\sugar}\\
        Benchmark
	&convert
	&ground
	&default
	&vsids
	&sat-pre
	&vsids/
	&ground
	&default
	&vsids
	&sat-pre
	&vsids/
	&convert
	&default
	&vsids\\
        &&&&&&sat-pre&&&&&sat-pre&&&\\
	\hline\hline
    1-fullins-5-5 & 2.02 & 1.41 & 13.96 & 11.28 & 5.15 & 3.66 & 0.91 & 10.72 & 12.15 & 6.72 & 7.18 & 1.73 & 2.40 & 2.19\\
3-fullins-5-6 & 3.55 & 32.39 & 17.07 & 14.90 & 21.75 & 6.52 & 11.26 & 13.08 & 16.52 & 11.53 & 14.02 & 5.36 & 1.91 & 1.50\\
4-fullins-4-7 & 2.20 & 4.45 & 28.35 & 45.59 & 22.02 & 28.38 & 2.29 & 20.03 & 39.92 & 28.60 & 42.91 & 2.80 & 2.19 & 4.81\\\HACK
abb313GPIA-7 & 5.02 & 46.97 & 0.67 & 1.85 & 0.71 & 2.01 & 20.33 & 31.75 & 2.27 & 41.61 & 1.63 & 7.18 & 6.05 & 0.05\\
abb313GPIA-8 & 5.14 & 51.23 & 411.17 & 460.56 & 521.11 & TO & 21.97 & 173.06 & 451.02 & 433.31 & 180.33 & 7.23 & 131.41 & 70.57\\
abb313GPIA-9 & 5.96 & 52.79 & TO & TO & TO & TO & 24.68 & 243.72 & TO & 289.54 & TO & 8.49 & 0.59 & 5.73\\\HACK
bibd-8-98-49-4-21\_glb & 3.68 & 28.30 & TO & 20.70 & 4.47 & 1.42 & 28.59 & 10.52 & 18.11 & 5.36 & 8.41 & 11.28 & 1.65 & 1.49\\
bibd-10-120-36-3-8\_glb & 5.20 & 69.94 & 46.68 & 11.62 & 8.21 & 1.13 & 76.65 & 15.84 & 4.55 & 2.47 & 7.93 & 13.63 & 7.28 & 0.85\\
bibd-25-25-9-9-3\_glb & 7.34 & 84.68 & TO & 434.23 & 3.50 & 1.13 & 17.53 & 235.11 & 177.27 & TO & 59.41 & 8.90 & 28.86 & 41.12\\
bibd-31-31-6-6-1\_glb & 9.74 & 254.13 & 159.32 & 3.34 & 156.71 & 14.64 & 39.84 & 12.24 & 0.18 & 0.17 & 0.11 & 17.39 & 77.95 & 0.20\\\HACK
C2-3-15 & 0.88 & 3.30 & 0.21 & 0.13 & 0.15 & 0.08 & 9.74 & 6.03 & 4.88 & 6.63 & 8.24 & 1.64 & 7.37 & 1.05\\
C4-1-61 & 1.44 & 24.44 & 4.66 & 2.86 & 4.90 & 2.80 & 148.33 & TO & TO & TO & TO & 3.40 & 3.06 & 9.87\\
C4-2-61 & 2.96 & 26.04 & 6.26 & 1.98 & 6.16 & 2.04 & 152.33 & 222.52 & 69.47 & 227.54 & 165.80 & 3.22 & 7.05 & 3.21\\
C5-3-91 & 2.73 & 71.55 & 31.42 & 10.32 & 31.87 & 11.91 & TO & TO & TO & TO & TO & 7.89 & 261.39 & TO\\\HACK
chnl-10-11 & 0.37 & 0.18 & 0.64 & 1.20 & 4.36 & 11.44 & 0.14 & 1.35 & 2.65 & 12.40 & 13.75 & 0.92 & 19.58 & 55.26\\
chnl-10-15 & 0.70 & 0.17 & 9.15 & 3.08 & 6.95 & 10.61 & 0.19 & 4.04 & 20.14 & 10.82 & 18.01 & 0.51 & 32.07 & 24.68\\
chnl-10-20 & 0.56 & 0.24 & 19.14 & 6.91 & 6.51 & 5.22 & 0.33 & 41.96 & 38.34 & 6.81 & 11.98 & 1.11 & 7.77 & 6.30\\
chnl10-15-pb-cnf-cr & 0.42 & 0.42 & 9.28 & 3.05 & 6.93 & 10.74 & 0.20 & 4.00 & 20.05 & 10.90 & 16.43 & 1.12 & 31.85 & 24.87\\\HACK
costasArray-14 & 0.46 & 0.82 & 3.54 & 2.93 & 3.65 & 4.87 & 0.62 & 18.63 & 18.19 & 24.79 & 6.52 & 2.03 & 0.25 & 0.07\\
costasArray-15 & 0.52 & 0.94 & 17.96 & 60.17 & 75.50 & 17.95 & 0.78 & 62.10 & 136.40 & 4.96 & 8.83 & 1.59 & 18.41 & 8.16\\
costasArray-16 & 0.52 & 1.20 & 15.13 & 30.18 & 69.69 & 0.99 & 1.03 & 130.75 & 203.94 & 245.02 & 36.31 & 1.71 & 32.65 & 33.25\\
costasArray-17 & 0.56 & 1.50 & TO & TO & TO & 232.64 & 1.28 & TO & TO & TO & TO & 2.08 & 148.55 & 553.54\\\HACK
fischer-1-2-fair & 1.44 & 304.21 & 11.21 & 92.81 & 11.42 & 18.79 & 28.50 & 23.81 & 7.81 & TO & TO & 3.42 & 0.97 & 0.02\\
fischer-2-3-fair & 4.18 & 348.84 & 0.85 & 21.11 & 0.79 & 6.50 & 68.42 & 23.19 & 3.69 & 23.71 & 3.54 & 7.19 & 0.13 & 0.04\\
fischer-3-8-fair & 30.18 & 515.69 & TO & 458.58 & TO & 60.32 & 376.33 & TO & TO & TO & TO & 35.16 & 22.05 & 37.27\\
fischer-4-6-fair & 35.12 & 535.93 & 282.84 & 242.05 & 386.37 & 274.27 & 384.19 & TO & TO & TO & TO & 39.15 & 17.16 & 8.79\\
fischer-6-1-fair & 4.40 & 457.55 & 0.91 & 12.16 & 0.90 & 14.68 & 150.78 & TO & 10.29 & TO & 5.46 & 13.20 & 1.79 & 2.69\\\HACK
magicSquare-6\_glb & 0.29 & 2.72 & 46.20 & 1.23 & 4.27 & 1.49 & 4.20 & 9.50 & 2.42 & 18.39 & 1.77 & 2.26 & 0.79 & 0.46\\
magicSquare-7\_glb & 0.28 & 8.04 & 7.54 & 78.17 & 1.15 & 30.18 & 15.63 & 51.08 & 9.59 & 51.00 & 7.41 & 3.00 & 4.10 & 2.20\\
magicSquare-8\_glb & 0.42 & 21.75 & TO & 318.69 & 477.11 & 47.22 & 53.61 & 331.09 & 19.14 & 124.79 & 36.60 & 6.29 & 5.50 & 3.72\\\HACK
mps-mzzv42z & 4.84 & 166.01 & 2.74 & 1.54 & 2.61 & 1.02 & 395.20 & 8.56 & 5.44 & 8.75 & 5.20 & 24.84 & 5.44 & 3.28\\
mps-p2756 & 1.74 & 278.02 & 4.98 & 11.89 & 5.04 & 5.29 & TO & TO & TO & TO & TO & 161.39 & 1.32 & 1.20\\
mps-red-air06 & 11.59 & 272.57 & 139.86 & 381.61 & 34.78 & 25.27 & TO & TO & TO & TO & TO & 27.73 & 0.63 & TO\\
mps-red-fiber & 1.28 & 92.78 & 5.07 & 3.78 & 5.34 & 5.39 & 388.43 & 33.55 & TO & 33.43 & TO & 38.74 & 2.05 & 6.76\\\HACK
queensKnights-50-5-add & 1.81 & 16.86 & 18.62 & 24.71 & 44.44 & 18.79 & 28.12 & 558.69 & 85.48 & 537.34 & 44.58 & 3.25 & 12.15 & 0.66\\
queensKnights-50-5-mul & 3.50 & 17.31 & 40.78 & 48.54 & 38.57 & 20.64 & 30.33 & 221.08 & 67.39 & TO & 50.76 & 3.29 & 2.10 & 0.63\\
queensKnights-80-5-mul & 2.47 & 77.52 & 181.70 & 354.74 & 172.33 & 383.73 & 126.32 & TO & 426.53 & TO & 414.97 & 5.81 & 7.61 & 2.51\\
queensKnights-100-5-add & 3.49 & 163.50 & TO & TO & TO & TO & 243.06 & TO & TO & TO & TO & 6.92 & 18.45 & 4.39\\\HACK
ramsey-16-3 & 1.07 & 0.51 & 1.17 & 1.16 & 0.18 & 0.10 & 0.39 & 6.74 & 0.53 & 0.78 & 1.04 & 2.19 & 1.99 & 112.63\\
ramsey-30-4 & 2.13 & 5.91 & 150.64 & 180.79 & 28.56 & 26.47 & 2.75 & 198.83 & 98.97 & 82.69 & 32.39 & 2.35 & 8.09 & 8.67\\
ramsey-33-4 & 2.53 & 7.58 & 405.05 & 279.25 & 93.36 & 66.78 & 4.05 & TO & 193.49 & 173.67 & 109.45 & 3.13 & 32.02 & 39.82\\
ramsey-34-4 & 2.34 & 8.46 & TO & TO & 366.41 & 109.45 & 4.14 & TO & TO & TO & 317.08 & 2.78 & 67.41 & 38.71\\\HACK
ruler-34-9-a4 & 1.15 & 13.01 & 21.70 & 29.62 & 19.05 & 19.16 & 3.08 & 35.29 & 39.71 & 38.59 & 39.19 & 4.30 & 41.98 & 41.16\\
ruler-44-10-a4 & 1.56 & 37.60 & 367.84 & 233.65 & 302.54 & 280.79 & 8.39 & 567.20 & 542.78 & TO & 483.85 & 9.91 & 405.96 & 446.66\\
ruler-44-9-a4 & 1.04 & 23.41 & 182.48 & 124.14 & 172.79 & 118.37 & 6.18 & 351.14 & 13.41 & 102.36 & 34.25 & 6.61 & TO & 352.28\\
ruler-55-10-a3 & 1.08 & 5.89 & TO & 500.84 & TO & TO & 3.19 & TO & TO & TO & TO & 1.73 & 43.23 & 70.22\\\HACK
super-jobShop-e0ddr1-8 & 1.21 & 6.42 & 555.34 & 19.21 & 14.32 & 3.67 & 3.66 & 6.10 & 0.11 & 6.92 & 1.05 & 1.16 & 1.21 & 0.49\\
super-jobShop-e0ddr2-1 & 1.08 & 7.59 & TO & 32.94 & 8.38 & 12.14 & 3.81 & 16.08 & 0.84 & 2.56 & 3.87 & 1.51 & 5.41 & 0.76\\
super-jobShop-enddr2-3 & 1.00 & 8.60 & TO & 116.06 & 76.55 & 12.93 & 3.95 & 13.21 & 2.59 & 5.30 & 6.33 & 2.15 & 1.19 & 0.72\\\HACK
super-os-taillard-7-4 & 1.02 & 35.84 & TO & 493.02 & TO & 454.02 & 15.35 & 36.02 & 36.35 & 35.06 & 32.14 & 2.08 & 1.14 & 0.89\\
super-os-taillard-7-6 & 1.01 & 35.51 & 115.70 & TO & 112.53 & TO & 14.49 & 169.99 & 30.25 & 144.26 & 25.89 & 2.14 & 4.20 & 0.76\\
super-os-taillard-7-7 & 0.84 & 32.73 & 155.31 & 270.49 & 132.71 & 311.67 & 13.53 & 23.55 & 17.77 & 25.74 & 15.78 & 1.95 & 0.80 & 0.78\\
super-os-taillard-7-8 & 0.95 & 33.23 & 431.15 & 279.09 & 475.51 & 328.50 & 14.46 & 46.95 & 25.64 & 40.83 & 24.06 & 2.01 & 1.19 & 0.96\\\HACK
zeroin-i-1-10 & 1.97 & 2.41 & 20.92 & 27.16 & 14.97 & 16.18 & 1.74 & 22.09 & 28.81 & 26.62 & 18.56 & 2.17 & 3.48 & 2.57\\
zeroin-i-3-10 & 1.76 & 1.92 & 7.24 & 33.14 & 11.43 & 16.93 & 1.52 & 13.45 & 29.43 & 19.72 & 31.67 & 2.22 & 4.57 & 4.11\\\HACK
ii-32c4 & 3.73 & 21.92 & 0.24 & 0.02 & 0.03 & 0.03 & 56.69 & 0.02 & 0.01 & 0.03 & 0.04 & 21.67 & 8.77 & 0.76\\
ii-32d3 & 2.80 & 11.87 & 4.58 & 13.70 & 0.39 & 0.26 & 25.12 & 3.54 & 0.31 & 0.54 & 1.35 & 12.33 & 71.17 & 0.20\\
p2756 & 2.06 & 269.82 & 3.87 & 12.55 & 3.99 & 6.13 & TO & TO & TO & TO & TO & 162.72 & 4.26 & 2.41\\
ooo-burch-dill-3-accl-ucl & 2.46 & 14.06 & 13.80 & 18.60 & 6.13 & 5.95 & 5.82 & 30.05 & 24.18 & 12.73 & 4.61 & 2.77 & 1.28 & 1.00\\
ooo-tag14 & 6.93 & 191.71 & 145.03 & 144.09 & 158.56 & 158.99 & 44.99 & 109.54 & 310.78 & 20.82 & 25.11 & 10.67 & 6.54 & 6.07\\\hline\hline
Average Time & 3.51 & 80.21 & 188.67 & 149.80 & 129.06 & 103.87 & 91.49 & 209.46 & 184.50 & 218.43 & 169.76 & 12.56 & 37.47 & 54.27\\
Timeouts & 0 & 0 & 12 & 5 & 6 & 5 & 4 & 14 & 13 & 17 & 13 & 0 & 1 & 2\\\hline

  \end{tabular}}
\end{table}

Table~\ref{tab:experiments} reports runtime results in seconds,
separated into conversion time of \aspartame\ from CSP instances
to ASP facts (first ``convert'' column) and of \sugar\ from CSP instances
to CNF, \gringo\ times for grounding ASP encodings relative to facts,
and finally columns for the search times of \clasp\ with the aforementioned
options.
Each computational phase was restricted to 600 seconds, and
timeouts counted in the last row of Table~\ref{tab:experiments} are taken
as 600 seconds in the second last row providing average runtimes.
Looking at these summary rows, we observe that our two ASP encodings
are solved most effectively when vsids decision heuristic and SAT preprocessing
are both enabled; unlike this, neither decision heuristic dominates the other on
CNF input.
Apparently, \clasp\ on CNFs generated by \sugar\ still has a significant edge on
facts by \aspartame\ combined with either ASP encoding.
In particular, we observe drastic performance discrepancies on some
instance families (especially ``fischer'' and ``queensKnights''), 
where \clasp\ performs stable on CNFs from \sugar\ but runs into trouble
on corresponding ASP instances.
Given that \aspartame\ and its ASP encodings are prototypes,
such behavior does not disprove the basic approach,
but rather motivates future investigations of the reasons for performance
discrepancies.
For one, we conjecture that normalizations of global constraints that are
not yet supported by \aspartame\ are primarily responsible for large
instance sizes and long search times on some instance families.
For another, we suppose that both of our ASP encodings are still quite
naive compared to years of expertise manifested in \sugar's CNF construction.
However, the observation that our dedicated ASP encoding has on edge the
SAT-inspired one and yields significant performance improvements on some
instance families (``C2-3-15''--``C5-3-91'' and ``mps'') clearly
encourages further investigations into ASP encodings of CSP instances.



\section{Related Work}
\label{sec:related:work}

Unlike 
approaches to constraint answer set solving,
e.g., \cite{drewal10a,geossc09a,balduccini09a,ostsch12a},
which aim at integrating CSP and ASP solving (engines),
the focus of \aspartame\ lies on pure CSP solving.
In fact, \aspartame's approach can be regarded as a first-order alternative to
SAT-based systems like \sugar~\cite{tatakiba09a},
where the performance of the underlying SAT solver is crucial.
However, it is now becoming recognized that the SAT encoding to be used
also plays an important role \cite{DBLP:series/faia/Prestwich09}.
There have been several proposals of encoding constraints to SAT:
direct encoding \cite{DBLP:conf/ijcai/Kleer89,DBLP:conf/cp/Walsh00},
support encoding \cite{DBLP:journals/ai/Kasif90,DBLP:conf/ecai/Gent02},
log encoding \cite{DBLP:conf/ifip/IwamaM94,DBLP:journals/dam/Gelder08},
log support encoding \cite{DBLP:conf/cp/Gavanelli07},
regular encoding \cite{DBLP:conf/sat/AnsoteguiM04},
order encoding \cite{crabak94a,tatakiba09a},
and compact order encoding \cite{DBLP:conf/sat/TanjoTB12}.

The order encoding, where Boolean variables represent
whether $x \le i$ holds for variables~$x$ and values~$i$,
showed good performance for a wide range of CSPs
\cite{crabak94a,mecost13a,DBLP:conf/sat/AnsoteguiM04,DBLP:conf/cp/BailleuxB03,MODREF04:GentN04,DBLP:journals/dam/InoueSUSBT06,sointabana10a,DBLP:journals/constraints/OhrimenkoSC09,BanbaraMTI:LPAR:2010}.
Especially, the SAT-based constraint solver \sugar\
became a winner in
global constraint categories at the 2008 and 2009 CSP solver competitions
\cite{DBLP:journals/constraints/LecoutreRD10}.
Moreover,
the SAT-based CSP solver BEE \cite{DBLP:journals/tplp/MetodiC12}
and the CLP system B-Prolog \cite{zhou2013a}
utilize the order encoding.
In fact, the order encoding provides a compact translation of
arithmetic constraints, while also maintaining bounds consistency by unit propagation.
Interestingly, it has been shown that the order encoding
is the only existing SAT encoding that can reduce
tractable CSP to tractable SAT \cite{DBLP:conf/sat/PetkeJ11}.




\section{Conclusion}\label{sec:discussion}

We presented an alternative approach to solving finite linear CSPs based on ASP.
The resulting system \aspartame\
relies on high-level ASP encodings and delegates both the grounding
and solving tasks to general-purpose ASP systems.
We have contrasted \aspartame\ with its SAT-based ancestor \sugar,
which delegates only the solving task to off-the-shelf SAT solvers, while using dedicated algorithms
for 
constraint preprocessing.
Although \aspartame\ does not fully match the performance of \sugar\ from a global perspective,
the picture is fragmented and leaves room for further improvements.
This is to say that different performances are observed on distinct classes of CSPs,
comprising different types of constraints.
Thus, it is an interesting topic of future research to devise more appropriate ASP encodings for
such settings.
Despite all this,
\aspartame\ demonstrates that ASP's general-purpose technology allows to compete with state-of-the-art
constraint solving techniques,
not to mention that \aspartame's intelligence is driven by an ASP encoding of
less than 100 code lines (for non-deterministic predicates subject to search).
In fact, the high-level approach of ASP facilitates extensions and variations of
first-order encodings for dealing with particular types of constraints.
In the future, we thus aim at more exhaustive investigations of
encoding variants, e.g., regarding alldifferent constraints,
as well as support for additional kinds of global constraints.

%


\paragraph{Acknowledgments}

This work was partially funded 
by 
the Japan Society for the Promotion of Science (JSPS)
under grant 
KAKENHI 24300007
as well as
the German Science Foundation (DFG) under grant 
SCHA 550/8-3 
and
SCHA 550/9-1.   
We are grateful to the anonymous reviewers for many helpful comments.




\end{document}